\documentclass[letterpaper, 10 pt, conference]{ieeeconf}  % Comment this line out for the final manuscript

\IEEEoverridecommandlockouts                              % This command is only needed if 
\usepackage{graphicx}
\graphicspath{{figures/}}
\usepackage{subfigure}
\usepackage{epstopdf}
\usepackage{booktabs} % For tables
\usepackage{amsfonts} % For blackboard math fonts

\newcommand{\reffig}[1]{Fig.~\ref{#1}}

\title{\LARGE \bf
CHARM: A Hierarchical Deep Learning Model for Classification of Complex Human Activities Using Motion Sensors
}

\author{Eric Rosen$^{1}$ and Doruk Senkal$^{1}$% <-this % stops a space
\thanks{\textit{(Corresponding author: Doruk Senkal.)}}
\thanks{$^{1}$Eric Rosen and Doruk Senkal are with Meta Platforms, Inc.,
	 1 Hacker Way, Menlo Park, CA 94025
	(e-mail: ericrosen@fb.com; dsenkal@fb.com)%
}
}

\begin{document}

\maketitle
\markboth{IEEE Robotics and Automation Letters. Preprint Version. Accepted Month, Year}
{FirstAuthorSurname \MakeLowercase{\textit{et al.}}: ShortTitle}

\begin{abstract}
	In this paper, we report a hierarchical deep learning model for classification of complex human activities using motion sensors. In contrast to traditional Human Activity Recognition (HAR) models used for event-based activity recognition, such as step counting, fall detection, and gesture identification, this new deep learning model, which we refer to as CHARM (Complex Human Activity Recognition Model), is aimed for recognition of high-level human activities that are composed of multiple different low-level activities in a non-deterministic sequence, such as meal preparation, house chores, and daily routines. CHARM not only quantitatively outperforms state-of-the-art supervised learning approaches for high-level activity recognition in terms of average accuracy and F1 scores, but also automatically learns to recognize low-level activities, such as manipulation gestures and locomotion modes, without any explicit labels for such activities. This opens new avenues for Human-Machine Interaction (HMI) modalities using wearable sensors, where the user can choose to associate an automated task with a high-level activity, such as controlling home automation (e.g., robotic vacuum cleaners, lights, and thermostats) or presenting contextually relevant information at the right time (e.g., reminders, status updates, and weather/news reports). In addition, the ability to learn low-level user activities when trained using only high-level activity labels may pave the way to semi-supervised learning of HAR tasks that are inherently difficult to label.

	%\vspace{0.67ex}
	%\textit{Index Terms}---Human Detection and Tracking, Intention Recognition, Deep Learning Methods, Human-Centered Automation.
\end{abstract}

%\begin{IEEEkeywords}
%	Human-Centered Automation, Human Detection and Tracking, Deep Learning Methods, Intention Recognition.
%\end{IEEEkeywords}
% Use for final version

\section{Introduction}
Human Activity Recognition (HAR), time-domain classification of human activities using sensor data, has been garnering increased interest with the advent of wearable sensing technologies. In today's wearable devices, HAR enables important downstream applications, such as Human-Machine Interaction (HMI), gesture recognition for user interaction, as well as health and wellness tracking. While a wide variety of different sensor modalities can be used for HAR, such as vision or audio, this work focuses on a subclass of HAR that uses body-worn Inertial Measurement Units (IMUs) \cite{wang2019deep}. IMUs are ubiquitous in today's mobile and wearable devices, as they provide a low-cost, low-power, self-contained, and privacy-focused sensing modality for both indoor and outdoor HAR.

\begin{figure}
	\centering
	\includegraphics[width=3.335in]{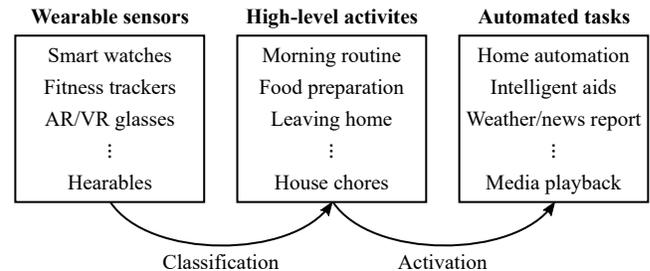}
	\caption{Examples of how classification of high-level activities can be used for Human-Machine Interaction (HMI). User can choose to activate a user-defined automated task, whenever a pre-determined high-level activity is performed.}
%	\caption{Potential examples of how classifying complex human activities can be useful for wearable devices. Users can specify desirable actions (e.g., play music, turn on lights) that are automatically activated when they perform high-level activities (e.g., room cleaning, morning routine). By high-level activities, we mean complex sequences of locomotion modes (e.g., walking, standing ) and manipulation gestures (e.g., grabbing, slicing) that can be sequenced in a variety of ways.}
	\label{fig:flow_diagram}
\end{figure}

Previous works on HAR using IMUs have largely focused on event-based activities, such as step counting, man-down detection, and gesture identification \cite{bulling2014tutorial}. Event-based activities, also termed low-level activities, are characterized by a discrete event that occurs within a finite time window, typically on the order of a few seconds or less. State-of-the-art approaches use machine learning techniques with hand-crafted features or neural networks to train supervised classification models to infer whether the event has occured within a specific time window. In this work, we investigate complex activities, also termed high-level activities, which contain multiple event-based activities in a non-deterministic sequence and occur over a long, highly variable duration (e.g., a person cleaning a room can manipulate and navigate the environment in highly varied ways and may take longer to complete the task for larger rooms). Related works in deep learning for computer vision have investigated recognizing high-level activities \cite{yang2009activity}, but high-level HAR using wearable sensors, such as IMUs, have not been studied extensively \cite{wang2019deep}. Our experimental results suggest that traditional machine learning approaches have challenges in high-level HAR in terms of consistent precision and recall across all classes. In addition, collecting supervised labels for low-level activities to support high-level activity recognition is costly and typically requires domain expertise. 

%However, detecting higher-level activities from motion sensor (such as non-periodic and non-sporadic events) have not been well-studied \citep{wang2019deep} because they can be highly irregular in execution and may take place over several minutes worth of time. 

We postulate that a scalable high-level HAR solution would need to autonomously extract relevant low-level motion representations to support high-level activity recognition without any supervised low-level motion labels.  We are motivated by the observation that high-level activities can be represented by a composition of low-level activities, similar to how a sentence is composed of individual words. For example, a high-level activity such as ``cleaning a room'' involves sequencing specific locomotion modes such as walking or standing and manipulation gestures such as picking up objects or opening drawers.

In this paper, we present CHARM (Complex Human Activity Recognition Model), a two-stage neural network architecture for classifying high-level activities from wearable sensor data that learns to represent low-level motion patterns without any low-level motion labels. The first stage uses a low-level neural encoder to compress short sequences of motion sensor data into a continuous feature representation. The second stage infers high-level activities from sequences of low-level encoder outputs strided across the raw data stream. CHARM exploits the natural structure of motion patterns during high-level activities, namely that they are complex compositions of low-level motion behaviors that may be sequenced in many ways. Our model architecture enables the low-level neural encoder to focus on shorter, localized motion patterns that characterize low-level motion patterns (over seconds) and the high-level encoder to focus on global patterns found across the long-horizon motion sensor stream (over minutes). 

We quantitatively test CHARM's capability to accurately classify high-level activities from motion sensor data by using the publicly available OPPORTUNITY data set \cite{roggen2010collecting,chavarriaga2013opportunity}, which includes four users performing four different high-level activities: ``morning routine'', ``coffee time'', ``lunch'', and ``cleanup''. To test if CHARM's low-level neural encoder learns semantically meaningful low-level motion representations when trained end-to-end for high-level activity recognition, we use Principal Component Analysis (PCA) to visualize low-level feature representations. We qualitatively find that the low-level neural encoder automatically learns to characterize relevant low-level motion patterns for high-level activity recognition without any explicit labels. This opens the door for new opportunities for efficiently learning to detect low-level activities on wearables by using labeled data of daily high-level activities. To our knowledge, this is the first demonstration of a deep neural network architecture classifying high-level human activities and automatically extracting low-level motion patterns using motion sensor data. The ability to detect high-level activities in a scalable manner may enable many new down-stream applications for Human-Machine Interaction, where the user can choose to tie a user-defined automated task to a high-level activity, \reffig{fig:flow_diagram}.%We find that CHARM's average classification accuracy on raw sensor data ($95\%$) outperforms an event-based multi-layer perceptron ($90\%$) and classical machine learning approaches with hand-crafted features (Random Forests (RF) ($0.86\%$), and Support Vector Machines (SVM) ($0.88$)). CHARM is also the only approach that has a minimum F1 score across classes above $0.90$, whereas all other approaches have a minimum F1 score below $0.75$.
%Temporal Convolutional Networks, a state-of-the-art baseline for HAR, classification accuracy by $10\%$. 

%This paper has three key contributions: 
%\begin{enumerate}
%	\item A novel hierarchical neural network architecture for supervised high-level activity recognition with motion sensor data, called CHARM (Complex Human Activity Recognition Model). CHARM employs two stages that are trained end-to-end: a low-level neural encoder that learns continuous feature representations of short-horizon time windows, and a high-level neural encoder that infers high-level activities from sequences of low-level encoder outputs.
%	\item A quantitative validation of CHARM's classification accuracy for high-level HAR using the publicly available OPPORTUNITY data set against multi-layer perceptrons, Random Forests (RF), and Support Vector Machines (SVM). To our knowledge, this is the first evaluation of these machine learning model's performance for high-level activity recognition on the OPPORTUNITY data set. 
%	\item A visual examination and discussion of CHARM's low-level neural encoder's ability to learn semantically meaningful representation of low-level motion patterns without supervised labels. We investigate both locomotion and manipulation gestures and provide insights and key lessons learned that may aid future research into high-level activity recognition with motion sensors.
%\end{enumerate}

\section{Related Work}
Human Activity Recognition (HAR) can be divided into two broad categories based on the sensing modality \cite{wang2019deep}: vision-based HAR and sensor-based HAR. Vision-based HAR focuses on using visual information such as images, whereas sensor-based HAR focus on other modalities like motion, audio, and biosensors located on the human body or objects in a smart home. Our review of related work focuses only on HAR using body-worn sensors, such as Inertial Measurement Units that consist of accelerometers and gyroscopes for sensing linear accelerations and angular velocities at the location they are worn.

In \cite{chen2015deep}, a Convolutional Neural Network (CNN) was used for recognizing human activities with accelerometer data and experimentally demonstrated that the deep learning architectures can achieve high accuracy for classification of locomotion modalities. In \cite{gjoreski2016comparing}, CNNs and other classical machine learning methods (Random Forests with hand-crafted features, naive bayes, and support vector machines) were used to perform HAR on elementary activities with wrist-based accelerometers. It was found that with a large enough data set, the deep learning approach outperformed the other models. Choosing an appropriate feature representation of the raw motion sensor data has been shown to be important for successful classification as demonstrated by methods leveraging hand-crafted features based on the sensor data and using bag-of-words approaches \cite{kheirkhahan2017bag, aslan2020human}.

An advantage of deep learning approaches is that learned feature representations from raw motion sensor data can be reused for classifying different activities and can be more robust to new on-body sensor locations via transfer learning \cite{morales2016deep}. These works have demonstrated the effectiveness of deep learning models for HAR with motion sensors, but have only focused on event-based activities, such as locomotion and event-based activity recognition. Other works have investigated recurrent models such as Long Short Term Memory (LSTM) networks for HAR \cite{edel2016binarized}, which are powerful inference models because they can operate on arbitray length sequences. However, \cite{bai2018empirical} found that simple convolutional network architecture outperforms canonical recurrent models on a wide variety of sequence modeling tasks, which inspired our particular instantiation of CHARM in our experiments.

Detecting complex activities from motion sensor data has also been recognized as an important area of research, with a large focus on leveraging low-level motion patterns or body components to aid in classification. For example, in \cite{ryoo2006recognition} a context-free grammar based representation was leveraged to decompose and classify complex activities by representing them as compositions of body-part and gesture layers. In \cite{liu2015action2activity}, frequent patterns from low-level actions were mined to construct intermediate representations for complex activity recognition. These works all assume that the requisite low-level components have already been labeled, in addition to the high-level activities. CHARM instead automatically learns to represent the low-level motion patterns by only using labels of high-level, complex activities.

%\cite{edel2016binarized}
%\cite{gjoreski2016comparing}
%\cite{morales2016deep}

\section{Problem Formulation}
We follow \cite{wang2019deep} in defining the HAR problem. We suppose that a user is performing an activity belonging to a predefined activity set $A$:

\begin{equation} \label{eq:A}
	A = \{A_{i}\}^{m}_{i=1}\,,
\end{equation}

\noindent where $m$ denotes the number of activity classes. The user produces a stream of sensor data $s$ that captures signals of the activity:

\begin{equation} \label{eq:s}
	s = \{d_{1},d_{2},\ldots,d_{t},\ldots,d_{n}\}\,,
\end{equation}

\noindent where $d_{t}$ represents the sensor data with dimension $q$ ($d_{t} \in \mathbb{R}^{q}$) collected at time $t$. 

A solution to the HAR problem is to define a model $F$ that predicts the activity $A$ based on the sensor stream~$s$, while the true sequence of activities being performed (ground truth) is denoted as $A^{*}$.

We assume access to batches of supervised data sets $D_{b}$ that consists of pairs of sensor data streams and ground truth high-level activity labels:

\begin{equation}
	D = \{s,A^{*}\}\,.
\end{equation}

The goal is to find a model that minimizes the discrepancy between the predicted classes and the ground truth sequence of activities. We can model this as an optimization problem by defining a loss function $\mathcal{L}$ that captures the discrepancy between $F(s)$ and $A^{*}$ as a real number: 

\begin{equation}
	\mathcal{L}(F(s),A^{*}) \in \mathbb{R}\,. 
\end{equation}

Given a supervised data set of sensor streams and ground truth activity labels, we search for an $F$ that minimizes the loss function $\mathcal{L}$. We defer to \cite{wang2019deep,bulling2014tutorial} for additional details of a conventional HAR pipeline.

\section{Proposed Algorithm}

\begin{figure}
	\centering
	\includegraphics[width=3.264in]{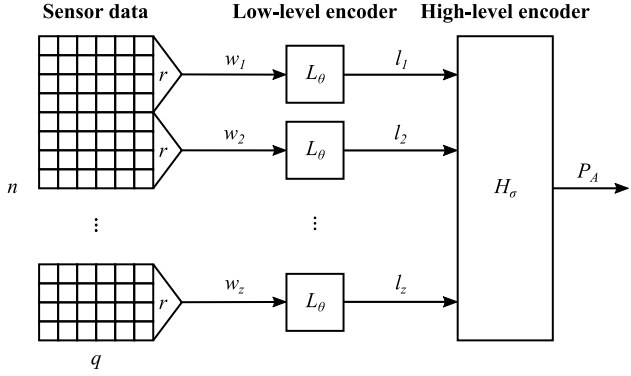}
	\caption{A schematic of the proposed deep learning architecture for high-level human activity recognition using wearable sensors, consisting of a low-level encoder to learn representations of short time-scale, event-based manipulation and locomotion activities present in the sensor data, which then feeds into a high-level encoder to make predictions about what high-level activity is occurring.}
	\label{fig:model}
\end{figure}

%Note that CHARM is not exposed to any labeled examples of low-level activities, and the low-level encoder and high-level encoder are only trained end-to-end on high-level activity labels.

To solve the high-level HAR problem with motion sensor data, we propose a hierarchical neural network architecture with two stages, which we term CHARM: a low-level neural encoder $L_{\theta}$ and a high-level neural encoder $H_{\sigma}$, where $\theta$ are the parameters to the low-level neural encoder $L$ and $\sigma$ are the parameters to the high-level encoder $H$ respectively. The CHARM network architecture uses the low-level encoder to repeatedly featurize sequential short time windows in the raw sensor data, and then uses the high-level neural encoder to classify the high-level activity based on the sequence of outputs from the low-level encoder, \reffig{fig:model}. By structuring the network to reuse the same low-level encoder across the input sensor data stream, the low-level encoder produces feature representations of the shorter motion patterns that are invariant to when they occur in the global sequence (similar to how a convolutional filter is invariant to translation in space via a limited kernel size). Note that the CHARM approach is agnostic to the particular deep neural network architecture of the low-level and high-level neural encoder. % (we provide details of the particular implementation for our experiments in Section \ref{CHARM}).

%\ernote{overloading certain symbols, change them up}
More formally, from the data set $D$, for any given sensor stream sample $s = \{d_{1},d_{2},\ldots,d_{t},\ldots,d_{n}\}$ with $n$ samples (the input $s$ is a tensor with shape $[n,q]$ and $d_{t}$ is a vector with shape $[q]$ for the $q$ motion stream channels), the first stage of the CHARM network applies a sliding window of size $r$ across the input steam with stride of equal length, producing a window sequence $s_{w} = \{w_{1},w_{2},\ldots,w_{t},\ldots,w_{z}\}$ ($s_{w}$ is a tensor with shape $[z,r,q]$ where $z$ is equal to $\frac{n}{r}$, and $w_{t}$ is a tensor with shape $[r,q]$). We then feed the batch of windowed frames $s_{w}$ into the low-level neural encoder $L_{\theta}(s_{w})$ to create a  $q'$-dimensional featurization of the batch of window frames $s_{l} = \{l_{1},l_{2},\ldots,l_{t},\ldots,l_{z}\}$ ($s_{l}$ is a tensor with shape $[z,q']$, and $l_{t}$ is a vector with shape $[q']$). The end of the first stage produces $s_{l}$ since it represents the featurization of all the windowed data frames. The second stage of CHARM takes the output of the first stage $s_{l}$, and feeds the input to the high-level neural encoder $H_{\sigma}$ which produces a normalized $m$-dimensional vector $P_{A}$ that represents the probability of each activity within the pre-determined activitiy set $A$ being present in the raw sensor stream~$s$.

When referring to the full CHARM approach, we use the notation $E_{\phi}$, where $\phi=\{\theta,\sigma\}$ refers to the collection of the low-level and high-level neural encoder parameters. During inference time, for a given sensor data stream $s$, the model $E_{\phi}$ outputs probability distribution $P_{A}$ for the pre-determined activitiy set $A$:

\begin{equation}
	P_{A}=E_{\phi}(s)\,.
\end{equation}

To estimate the parameters $\phi$, we perform stochastic gradient descent with batches of data $D_{b}$. We use the negative log-likelihood as our objective function.

Before training and validation, we apply standard preprocessing steps to the raw sensor data that are agnostic to the specific task and data set. We normalize the data by subtracting off the mean and dividing by the standard deviation for each of the sensor channels separately. If there are any sequences that have multiple simultaneous high-level labels present in a given data sequence, we do not use that sequence for training or validation.

\begin{figure*}\centering
	\subfigure[``Morning routine'' class]{\label{d}\includegraphics[width=3.4in]{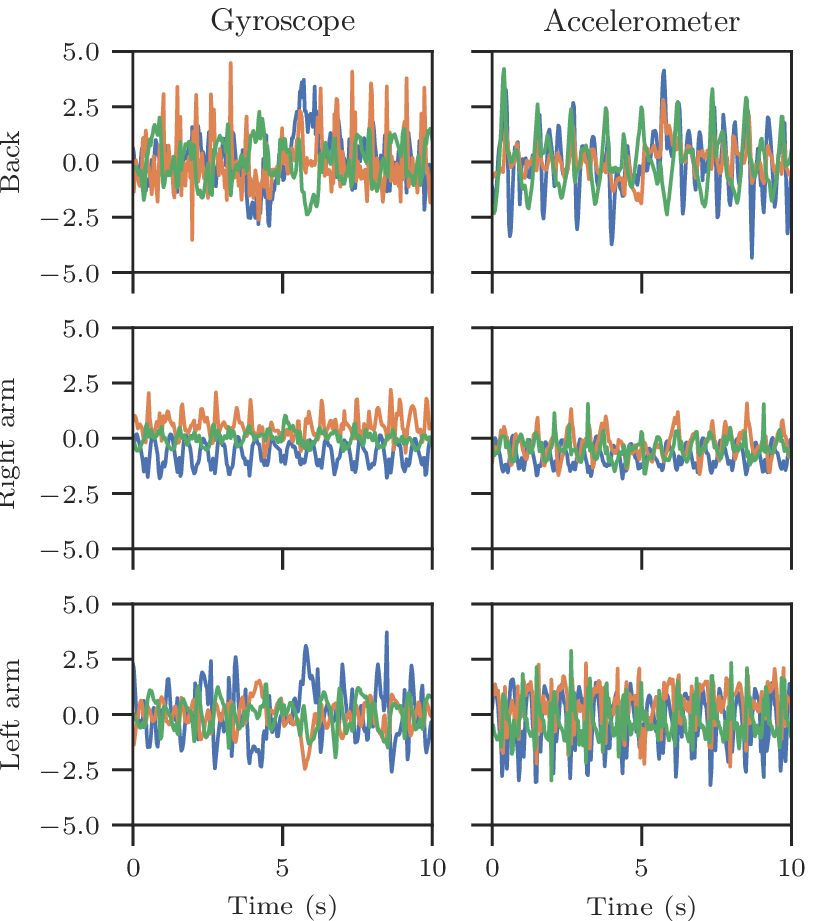}}	
	\hspace*{\fill}
	\subfigure[``Coffee time'' class]{\label{b}\includegraphics[width=3.4in]{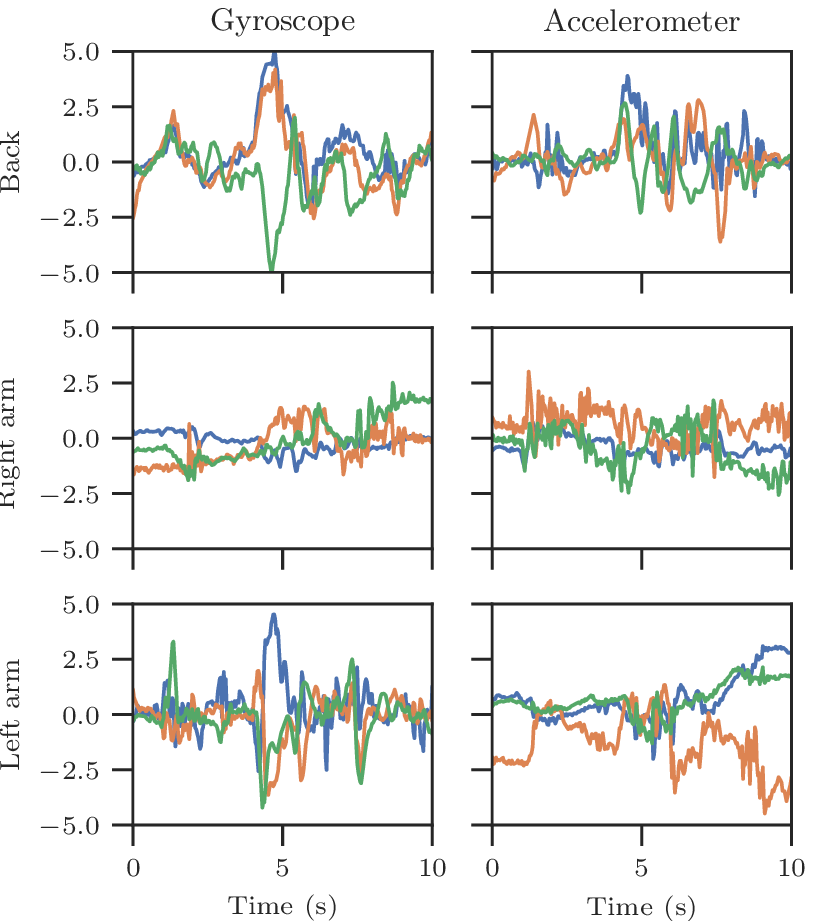}}	
	\subfigure[``Lunch'' class]{\label{c}\includegraphics[width=3.4in]{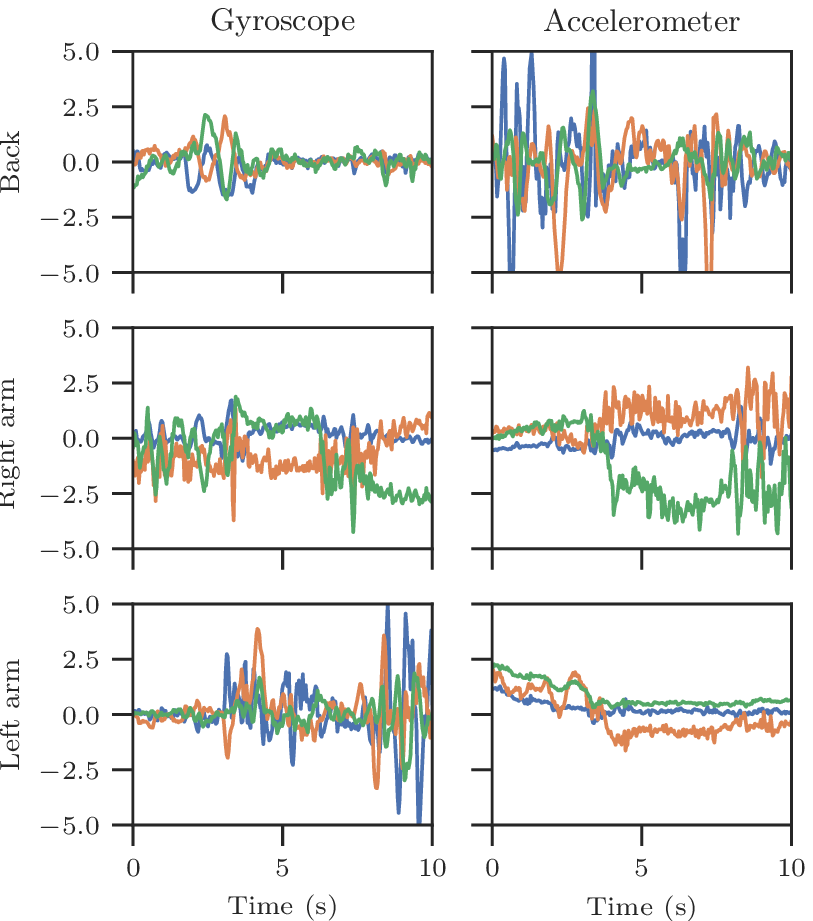}}
	\hspace*{\fill}
	\subfigure[``Cleanup'' class]{\label{a}\includegraphics[width=3.4in]{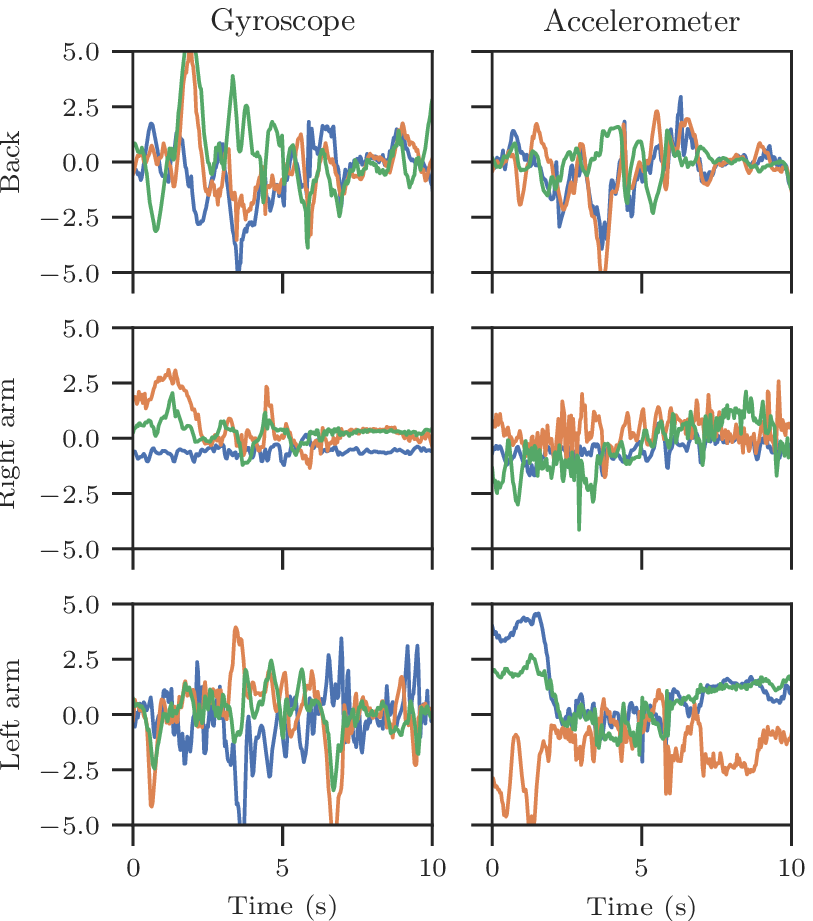}}	
	\caption{Raw motion sensor data from the four different high-level activity classes. Each visualization shows 10 s of data from the three sensor locations (Left arm, Right arm, and Back) for the triaxal gyroscope (rad/s) and accelerometer (gee) output. X, Y, and Z sensor axes are colored blue, orange, and green respectively.} 
	\label{fig:raw_sensor_data}
\end{figure*}

\begin{table}
	\centering
	\caption{\label{tab:activities} The high-level and low-level activities in the OPPORTUNITY data set that were used in the experiments. Note that CHARM is trained end-to-end only on the high-level activity labels.}
	\begin{tabular}{ccc}
		\toprule
		High-level activities & \multicolumn{2}{c}{Low-level activities}\\
		\cmidrule(lr){2-3}
		& Locomotion & Manipulation\\
		\midrule
		Morning routine  & Standing & Stir, Sip\\
		Coffee time & Walking & Cut, Spread, Bite\\
		Lunch & Sitting & Open, Close\\
		Cleanup & Lying & Lock, Unlock\\
		\bottomrule
	\end{tabular}
\end{table}

\section{Experimental Results}
\subsection{Data Set}
\label{OPPORTUNITY}
For the experiments we used the OPPORTUNITY data set, which is a publicly available data set for evaluating HAR models \cite{roggen2010collecting,chavarriaga2013opportunity}. The data set contains readings of motion sensor data from four users performing daily living activities in a room simulating an apartment flat, Table~\ref{tab:activities}. To collect the motion sensor data, participants wore a jacket outfitted with triaxal IMUs (30 Hz data sampling rate) located at lower left arm, lower right arm, and upper back positions, \reffig{fig:raw_sensor_data}. Participants were instructed to perform $5$ daily activities: 

\textbf{Morning routine}: Users began lying on a canvas chair. At the user's own pace, they stood up and went out the door of the apartment for a walk. After a leisurely walk, they then re-entered the flat and closed the door behind them.

\textbf{Coffee time}: The users prepared and drank a cup of coffee. Users grabbed a cup located on a rack, put coffee mix and milk into a machine, stirred the ingredients together, and drank the coffee at their leisure.

\textbf{Lunch}: Users prepared a sandwich and ate it. Sandwich preparation involved cutting two slices of bread with a knife and spreading cheese on the slices with a knife. The users then put salami on the slices, and the sandwiches were then placed onto a plate for eating.

\textbf{Cleanup}: After eating and drinking, users cleaned up their apartment. Users put food back to the original locations, put any dishes into the dishwasher, and wiped the tables with a towel.

\textbf{Relaxing}: Users walked around the apartment at their leisure. The users turned off appliances in the room, and then laid down back onto their chair to rest.

Although not explicitly instructed to, during the execution of these daily activities, the users performed atomic manipulation and locomotion actions, e.g., reaching for drawers, walking etc. To model the temporal decomposition of these daily-activities, the OPPORTUNITY data set contains three relevant label tracks aligned with the sensor data: a daily activity track, a locomotion activity track, and a manipulation activity track.
% The locomotion activity track consists of four modes: ``sitting'', ``lying'', ``walking'', and ``standing''. The manipulation track contains $9$ relevant manipulation behaviors for both the left and right arm: ``unlock'', ``lock'', ``open'', ``close'', ``sip'', ``bite'', ``cut'', ``stir'', and ``spread''. In general, during a single segment of a high-level activity, the user performed a sequence of manipulation and locomotion behaviors.
In this paper, we treat the daily activity labels as examples of high-level activities, while treating manipulation and locomotion labels as examples of low-level activities. 

\begin{figure}[!t]
	\centering
	\includegraphics[width=3.4in]{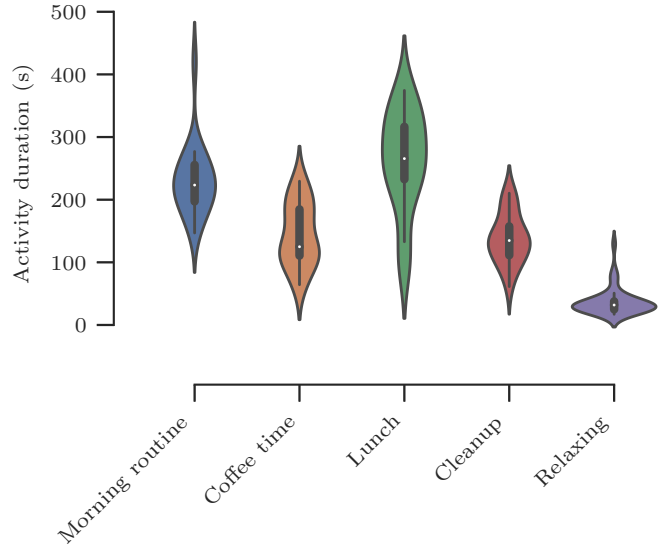}
	\caption{Violin plots showing the duration of high-level activities in the OPPORTUNITY data set. Note that the ``Relaxing'' class is unique in how short in length the activities are (typically less than ~100 s) compared to all the other activities (which range from ~100 to ~400 s).}
	\label{fig:durations}
\end{figure}

For our experiments, we leveraged the triaxal gyroscope and accelerometer data collected at the lower-left arm, lower-right arm, and upper back positions. This is because they are the most representative locations for existing wearable technology, e.g., wrist-based and head-mounted devices. Note that because of this, for all of our experiments, each data point in the sensor sequence had $18$ channels. 

Two major challenges in using the OPPORTUNITY data set for this study were the semantics of the null class and class imbalance: in the OPPORTUNITY data set sequences labeled as ``null'' had no interpretable meaning to them as they occured between labeled high-level activities and therefore contained motion patterns from multiple different high-level activities, making it a poor class to train and evaluate on. %represents around \~80\% of the class labels \cite{gjoreski2016comparing}.
For this reason, we do not include the ``null'' class in our experiments. In addition, for the high-level activity ``relaxing'', we found there was a significant imbalance in the activity length compared to all the other high-level activities. Specifically, the  longest instances of the ``relaxing'' activity were around $100$ s, while the shortest length instances of all the other high-level activities were around $100$ s and lasted up to several minutes, \reffig{fig:durations}. We therefore did not use the ``relaxing'' class, and only focused on learning from the four high-level other activity labels (``Morning routine'', ``Coffee time'', ``Lunch'', and ``Cleanup''). Note that we train our high-level activity recognition model only using these high-level labels, i.e., CHARM is never exposed to the low-level locomotion and manipulation labels included in the OPPORTUNITY data set. %We demonstrate in Section \ref{low-level-semantic-representations} how CHARM's low-level encoder, when trained end-to-end for high-level activity recognition, learns distinct feature representations of both locomotion and manipulation gestures without any supervised labels.

\begin{table*}
	\centering
	\caption{\label{tab:model_scores}  Precision (P), Recall (R) and F1 scores for classification of high-level activities using SVM, Random Forest (RF), Multi-layer Perceptron (MLP), and CHARM.}% Bold numbers represent the best performance across the models for each metric and class.}
\begin{tabular}{lccccccccccccccc}\toprule
	& \multicolumn{3}{c}{Coffee time} & \multicolumn{3}{c}{Morning routine} & \multicolumn{3}{c}{Cleanup} & \multicolumn{3}{c}{Lunch} & \multicolumn{3}{c}{Class average} 
	\\\cmidrule(lr){2-4}\cmidrule(lr){5-7}\cmidrule(lr){8-10}\cmidrule(lr){11-13}\cmidrule(lr){14-16}
	& P  & R & F1    & P  & R & F1  & P  & R & F1 & P  & R & F1 & P  & R & F1\\\midrule
	SVM & \textbf{1.00} & 0.91 & \textbf{0.95} & 0.82 & \textbf{1.00} & 0.90 & \textbf{1.00} & 0.59 & 0.74 & 0.88 & \textbf{1.00} & 0.94& 0.93 & 0.88 & 0.88 \\
	RF & 0.52 & 0.26 & 0.34 & 0.65 & \textbf{1.00} & 0.79 & 0.99 & 0.43 & 0.60 & 0.84 & 0.95 & 0.89 & 0.75 & 0.66 & 0.66 \\
	MLP & 0.75 & 0.62 & 0.68 & 0.88 & \textbf{1.00} & 0.94 & 0.91 & \textbf{0.92} & 0.91 & \textbf{0.99} & 0.93 & \textbf{0.96} & 0.88 & 0.87 & 0.87\\
	CHARM & 0.91 & \textbf{0.93} & 0.92 & \textbf{0.93} & \textbf{1.00} & \textbf{0.96} & 0.97 & 0.88 & \textbf{0.93} & 0.96 & 0.97 & \textbf{0.96} & \textbf{0.94} & \textbf{0.95} & \textbf{0.94}\\
	\bottomrule
	%e   & 11034 & 1.3e-7 & 3.9 & 15846 & 2.7e-11 & 5.6 \\
	%f & 21952 & 1.3e-7 & 6.2 & 31516 & 2.7e-11 & 8.8 \\
	%g & 15883 & 5.2e-8 & 7.1 & 32023 & 1.1e-11 & 1.4e1\\
	%h   & 11180 & 8.0e-9 & 4.3 & 17348 & 1.5e-11 & 6.6 
	
\end{tabular}
\end{table*}

\subsection{Models}
For our experiments, we evaluated $4$ different machine learning approaches: Support Vector Machines (SVMs), Random Forests, a Multi-Layer Perceptron, and CHARM. For SVMs and Random Forests, we evaluated the models with both raw sensor data and with hand-crafted features common for event-based activities. For these approaches, we applied $5$ standard hand-crafted features from \cite{zhang2011feature} to each of channels in the raw sensor data (note that because there are $18$ channels, the size of the resulting input vector dimension is $90$). The $5$ hand-crafted features we compute for each channels are: the mean, the variance, the difference between the maximum and minimum peaks in the signal, the minimum value, and the maximum value. We also performed hyperparameter sweeps for all models and reported the best results. The following subsections describe the hyperparameters for the models. For all the models, we train using data files of $3$ of the $4$ users from the OPPORTUNITY data set, and validate using the last held-out user. All approaches had input sequences with 2560 data points (at 30 Hz, ~85 s), each data point with $18$ channels consisting of $3$ sensor locations, each with $2$ sensor modalities (gyroscope and accelerometer) and $3$ sensor axes (X, Y, Z).

%\ernote{write these values in terms of math terms defined above}
\subsubsection{CHARM}
\label{CHARM}
The low-level encoder window size was $16$ (about 500 ms), with a two-layer fully-connected network with a hidden-state size of $32$ dimensions. The non-linear activation function between the layers was a Leaky ReLu with a negative slope of 0.01. The high-level encoder was a two-layer fully-connected network with an input sequence size of 160 data points, each with $32$ dimensions from the low-level encoder. We applied a softmax to the final output from the high-level encoder to calculate a probability distribution over the high-level classes. We used a cross-entropy loss as our optimization criterion, and gave weights to each class inversely proportional to the number of labels of each class in the training data. We trained with a batch size of 1 for 10 epochs. We use the Adam optimizer with a learning rate of $5e^{-4}$. We performed dropout during training between all the layers with a probability of $5\%$. At test time, we selected the class with the highest probability.

\subsubsection{Multi-Layer Perceptron (MLP)}
The MLP had $4$ fully-connected layers, each with a hidden size of $16$. The non-linear activation function between the layers was a Leaky ReLu with a negative slope of 0.01. We applied a softmax to the output from the final layer to calculate a probability distribution over the high-level classes.  We used a cross-entropy loss as our optimization criterion, and gave weights to each class inversely proportional to the number of labels of each class in the training data. We trained with a batch size of 1 for 10 epochs. We used the Adam optimizer with a learning rate of $5e^{-4}$ and performed dropout during training between all the layers with a probability of $5\%$.

\subsubsection{Support Vector Machines (SVMs)}
We found that Support Vector Machine performed better on the raw data rather than using the hand-crafted features. We used a radial-basis function as the kernel for the SVM, with a scaled gamma kernel coefficient. 

\subsubsection{Random Forests (RF)}
For the Random Forests, we apply the hand-crafted features to the input sequence. We use 100 trees in the forest for estimation, and we use the Gini criterion for measuring quality of splits in the tree. Splits were made in the tree until the leaves only contained points of the same class. 

\subsection{Classification Performance}
For each model, we report the precision, recall and F1 scores for all the high-level activity classes in Table~\ref{tab:model_scores}. For the precision, recall, and F1 score for each class, we bold the numbers of the model that performed best in that category. Overall, CHARM has the best performance across most of the metrics, with the multi-layer perceptron having the second-best performance, while the SVM and RF performed the worst, which was expected since the neural network approaches are able to learn nonlinear feature representations of the input data.  
%CHARM has the highest F1 score for three of the four classes (``Morning routine'' ($0.96$), ``Room cleaning'' ($0.93$), and ``Lunch preparation'' ($0.96$)), and has the second highest F1 score for the other class (``Coffee preparation'' ($0.92$)). CHARM also has the highest recall in two of the four classes (``Coffee preparation'' ($0.93$) and ``Morning routine'' ($1.00$)), and CHARM also has the highest precision for one of the four classes ``Morning routine'' ($0.93$)). 
The overall average accuracy of CHARM is also the highest of all the models ($0.95$). Almost all of CHARM's precision, recall, and F1 scores were above $0.90$ for all the classes, unlike the other models which do not perform as well and struggle to get a F1 score across all classes above $0.75$. This suggests that CHARM is best suited for the task of high-level activity recognition because it does not over-fit to any of motion patterns in the individual classes but also learns to effectively generalize from the training data to the validation set. 

\begin{figure*}
	\centering
	\subfigure[Food related actions]{\label{fig:manipulation_food_drink}\includegraphics[width=2.2in]{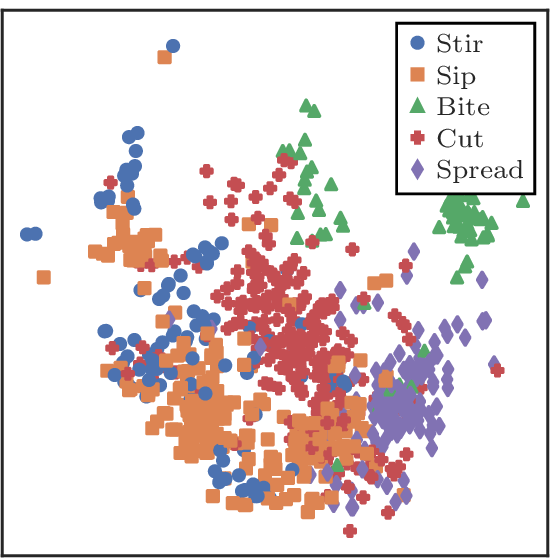}}
	\hspace*{\fill}
	\subfigure[Food-related vs object-related actions]{\label{fig:manipulation_object_cook}\includegraphics[width=2.2in]{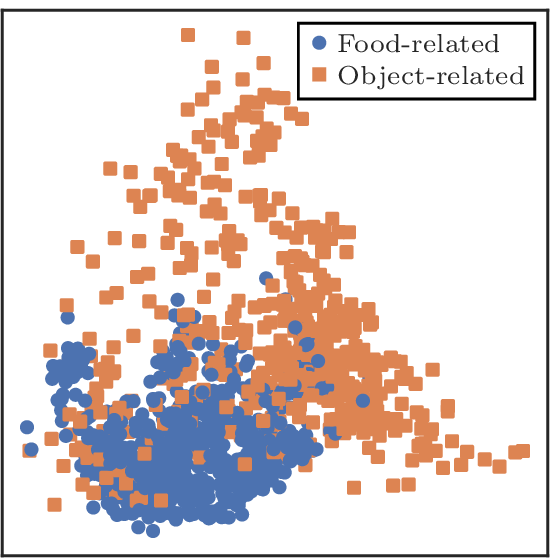}}		\hspace*{\fill}
	\subfigure[Locomotion actions]{\label{fig:locomotion}\includegraphics[width=2.2in]{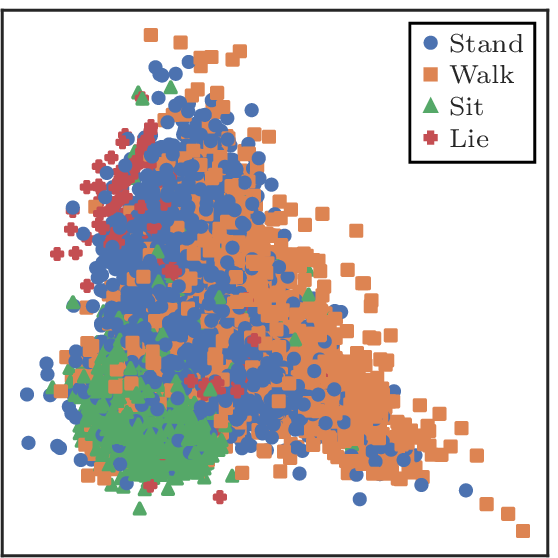}}
	\hspace*{0.25in}
	\caption{\mbox{2-D} visualization of low-level neural encoder representations for different low-level manipulation and locomotion activities. Even though the low-level encoder was not trained with any of these labels, the low-level neural encoder still learns to cluster related motion patterns. (a) Manipulation activities: stirring, sipping, biting, cutting, and spreading. (b) Food-related activities (stirring, sipping, biting, cutting, and spreading) vs object-related activities (unlocking, locking, opening, closing). (c) Locomotion modes: standing, walking, sitting, and lying down.}
	\label{fig:lowlevviz}
\end{figure*}

\subsection{Low-Level Semantic Representations}
\label{low-level-semantic-representations}
CHARM was inspired by the notion that a high-level activity consists of potentially repeated discrete event-based low-level motions that may be sequenced in many different ways to accomplish the task. Because we do not assume access to any labeled examples of low-level motions at training time, we are training both stages of the model end-to-end based only on high-level activity labels. 

Convolutional filters in computer vision have been known to learn semantically meaningful hierarchical feature representations of concepts in images \cite{zhang2018visual} when trained end-to-end for a downstream task. For example, when trained to detect objects and animals, visualizations of the feature representations in earlier layers show kernels that represent edge detectors, and later layers learn to detect shapes based on those edges. This inspired us to investigate whether CHARM's low-level encoder also learns representations of semantically meaningful concepts in the motion patterns present in the high-level activity sequences. 

To this end, we first trained the low-level and high-level encoder end-to-end on the high-level activity recognition tasks described in Section~\ref{OPPORTUNITY}. Then, we used the labeled data set for locomotion and manipulation activity tracks to generate labeled sequences with the same size as the low-level encoder (500 ms) for all of the low-level classes. We then fed all of the sequences into CHARM's low-level encoder and fed the high-dimensional output through a Principal Component Analysis (PCA) to reduce the feature representation to 2 dimensions. This dimensionality reduction technique enables us to visualize the learned representations of the low-level encoder using a \mbox{2-D} graph, where we can color the points based on the known label of the manipulation or locomotion sequence. We hypothesized that the low-level neural encoder would learn to cluster low-level motion patterns of the same class, even though the low-level encoder was never exposed to these labels. In \reffig{fig:lowlevviz}, we present visualizations for different manipulation and locomotion classes.

In \reffig{fig:manipulation_food_drink}, we plot the learned representations of $5$ manipulation behaviors: stirring, sipping, biting, cutting, and spreading. We qualitatively see that each of these manipulation behaviors have distinct clusters, which indicate that CHARM has learned to make distinct representations of these low-level motion patterns. Although ``biting'' and ``sipping'' are motion patterns that are very similar to each other in the raw motion sensor data (since both involve the user bringing their hand near their face), CHARM learns to separate these representations from each other and instead embed them closer to manipulation behaviors that are closely related to each other for classification of high-level activities. In addition, we notice that the ``biting'' activity is close to the ``spreading'' and ``cutting'' activities, whereas the ``sipping'' activity is close to the ``stirring'' activity. In practice, the ``biting', ``spreading'' and ``cutting'' activities almost always occur in the ``Lunch'' high-level activity, whereas ``sipping'' and ``stirring'' activities are almost exclusively performed during coffee preparation. We therefore hypothesize that the distance between embedded motion patterns inside CHARM's low-level encoder are related to how correlated those motion patterns are to the high-level activities, and not how similar the motion patterns are to each other in raw sensor data. In other words, CHARM's low-level encoder embeds motion patterns close to each other based on how relevant they are for classification of high-level activities, and not just based on how similar the underlying motion sensor data is.

To determine if this behavior was consistent across other manipulation behaviors, we investigated two classes of low-level activities: ``object-related'' activities and ``food-related'' activities, \reffig{fig:manipulation_object_cook}. Object-related activities were typically performed in ``cleanup'' and ``morning routine'' high-level activities, and included manipulation gestures like ``unlocking'', ``locking'', ``opening'', and ``closing''. On the other hand, food-related activities were mostly used in the ``lunch'' and ``coffee preperation'' classes, and involved manipulation behaviors like ``sipping'', ``cutting'', ``stirring'', ``bitting'', and ``spreading''. We plot the learned representations of these $2$ classes in \reffig{fig:manipulation_object_cook}, and qualitatively see that there are two distinct clusters of these classes. This further supports our hypothesis that CHARM's low-level encoder has learned to make distinct feature representations of low-level manipulation gestures, and that the distance between the embedded representations is proportional to how correlated the gestures are to the same high-level activities. In other words, low-level gestures that are prevalent in the same high-level activities are closer in CHARM's learned embedding space.

We also wanted to investigate whether CHARM learned distinct feature representations of locomotion motion patterns as well. In \reffig{fig:locomotion}, we plot the learned representations of $4$ locomotion behaviors: ``standing'', ``walking'', ``sitting'', and ``lying-down''. Similar to the manipulation gestures, we see that each of these locomotion modes has a distinct cluster, although there is much higher-overlap, especially between the ``sitting'' and ``standing'' class. However, this is expected because we only have access to motion sensor data, which makes distinguishing standing and sitting extremely challenging. In addition, ``lying-down'' and``standing'' have extremely high overlap since users who were lying-down during the experiments eventually had to stand up in order to do any of the high-level activities, and ``standing'' and ``walking'' have high overlap since users often walked and stood at locations when doing any of the high-level activities. Overall, since almost all of the high-level activities involved almost all of the locomotion modes except for ``lying-down'', which was almost exclusively done in the ``Morning routine'' class because users started lying down, most of the locomotion modes are closer in embedding space than the manipulation behaviors, but there still exist distinct clusters of these manipulation behaviors.

Overall, we postulate that these qualitative assessments validate that CHARM's low-level encoder is able to learn semantically meaningful representations of low-level motion patterns, even when it is only trained end-to-end on high-level activity labels. The important caveat to this is that, although CHARM learns to generally distinguish individual low-level motion patterns from each other, the distance between clusters is more closely related to how correlated low-level motion pattern are to similar high-level activities, and not the underlying low-level motion patterns themselves. This is important because it implies that the choice of high-level activity label and how common certain low-level motion patterns are in each high-level activity class have a large impact on what kinds of low-level motion patterns can be automatically distinguished. This ultimately makes sense because the model is trained end-to-end on high-level activity labels and feature representations are optimized so as to maximize performance on detecting high-level activities, so low-level motion patterns that are present in the same high-level activity will become closer in the embedding space because they are useful signals for determining what high-level activity is occurring. In other words, when determining, for example, whether a user is preparing lunch, it is sufficient to just check if they are either biting, spreading, or cutting, since as long as one of these activities is occurring it is highly likely that the user is preparing lunch since none of these manipulation behaviors occur in the other high-level activities. Therefore, the model learns to project these motion patterns to similar locations in the embedding space. 

\section{Conclusion}
In this paper, we investigated feasibility of using wearable sensor data and deep learning algorithms for the classification of high-level activity recognition.
%Whereas event-based activities are short and consistent in duration and have deterministic motion patterns, high-level activities are complex sequences of low-level behaviors that can be executed in many different ways and over long periods of time. 
Motivated by the observation that high-level activities can be represented by a composition of low-level activities, we developed a hierarchical deep learning architecture for HAR, %that is composed of two stages: a low-level neural encoder that operates on short-horizon time windows, and a high-level neural encoder 
which infers high-level activities from low-level encoder outputs.
% CHARM not only enables state-of-the-art classification performance on HAR for high-level activities, but when CHARM is trained end-to-end, it enables the low-level encoder to learn representations of distinct manipulation and locomotion behaviors without an explicit labels, which we verify by plotting low-dimensional visualizations of the low-level encoders output. 

Our results suggest that representations of semantically meaningful manipulation and locomotion behaviors can automatically be learned from end-to-end training of deep neural network models for high-level activity recognition using wearable sensor data to enable new applications in Human-Machine Interaction, such as intention recognition and human-centered automation, as well as health and wellness tracking. New future research directions may include: using automatically learned low-level feature representations for improving event-based detection via transfer learning and application of CHARM architecture for classification of other high-level activities.

\bibliography{charm_bibliography}
\bibliographystyle{IEEEtran}

\end{document}